\documentclass[journal]{IEEEtran}
\usepackage{amsmath}
\usepackage{url}
\usepackage{graphicx}
\usepackage{epstopdf}
\usepackage{subfigure}
\usepackage{hyperref}
\usepackage{multirow}
\usepackage{booktabs}
\usepackage{amssymb}

\ifCLASSINFOpdf
\else
\fi
\begin{document}
\title{Mixed-Precision Quantized Neural Networks with Progressively Decreasing Bitwidth for Image Classification and Object Detection}

\author{Tianshu Chu, Qin Luo, Jie Yang, Xiaolin Huang, \IEEEmembership{Senior Member, IEEE}
	
\thanks{This work was jointly supported by the National Natural Science Foundation of China (61603248) and the Science and Technology Project of State Grid Corporation (Research and Demonstration Application of Monitoring and Management Technology of City Energy System Based on Large Data and Artificial Intelligence CEGHJS1800002).}

\thanks{T. Chu, Q. Luo, J. Yang and X. Huang are with the Institute of Image Processing and Patter Recognition, Shanghai Jiao Tong University, Shanghai 200240, China (e-mail: chutianshu@sjtu.edu.cn; tomqin@sjtu.edu.cn; jieyang@sjtu.edu.cn; xiaolinhuang@sjtu.edu.cn). }
}


\markboth{Journal of \LaTeX\ Class Files,~Vol.~14, No.~8, August~2015}%
{Shell \MakeLowercase{\textit{et al.}}: Bare Demo of IEEEtran.cls for IEEE Journals}



\maketitle

\begin{abstract}
	
	Efficient model inference is an important and practical issue in the deployment of deep neural network on resource constraint platforms. Network quantization addresses this problem effectively by leveraging low-bit representation and arithmetic that could be conducted on dedicated embedded systems. 
	In the previous works, the parameter bitwidth is set homogeneously and there is a trade-off between superior performance and aggressive compression. Actually the stacked network layers, which are generally regarded as hierarchical feature extractors, contribute diversely to the overall performance. For a well-trained neural network, the feature distributions of different categories differentiate gradually as the network propagates forward. Hence the capability requirement on the subsequent feature extractors is reduced. It indicates that the neurons in posterior layers could be assigned with lower bitwidth for quantized neural networks. Based on this observation, a simple but effective  mixed-precision quantized neural network with progressively decreasing bitwidth is proposed to improve the trade-off between accuracy and compression. 
	Extensive experiments on typical network architectures and benchmark datasets demonstrate that the proposed method could achieve better or comparable results while reducing the memory space for quantized parameters by more than 30\% in comparison with the homogeneous counterparts. In addition, the results also demonstrate that the higher-precision bottom layers could boost the 1-bit network performance appreciably due to a better preservation of the original image information while the lower-precision posterior layers contribute to the regularization of $k-$bit networks.
	
\end{abstract}


% \fi
\IEEEpeerreviewmaketitle

\section{Introduction}

\IEEEPARstart{T}{he} deep convolutional neural networks (CNNs) have achieved state-of-the-art results on computer vision tasks, such as image categorization \cite{krizhevsky2012alexnet} \cite{simonyan2015vgg} \cite{he2016ResNet} \cite{huang2017densely}, object detection \cite{ren2015rcnn} \cite{ren2017faster} and semantic segmentation \cite{long2015fcn} \cite{he2017mask}. These achievements depend on the extreme model complexity that overfits the distribution of numerous training data. However, this also leads to a large over-parameterized model and dramatical computation cost. A typical CNN often takes hundreds of MB memory space, i.e. 170MB for ResNet-101 \cite{he2016ResNet}, 250MB for AlexNet \cite{krizhevsky2012alexnet}, 550MB for VGG-19 \cite{simonyan2015vgg} and requires billions of FLOPs per image during inference that rely on powerful GPUs. This challenges the deployment of CNNs on the edge devices such like mobile phones and drones. Thus the network compression and acceleration are an important issue in deep learning research and application.  

Several techniques have been proposed to tackle this issue via compact neural architecture design \cite{zoph2018nasnet}, model pruning \cite{frankle2018lottery} and network quantization \cite{hubara2017qnn}.
While the network topology remaining unchanged, the quantization is able to reduce the model size greatly to only a fraction of the origin by utilizing low-precision representation of parameters \cite{courbariaux2015binaryconnect}. Further more, the internal features are also could be quantized. Then the model inference is accelerated significantly by convert the expensive float-point arithmetic into more effective fixed-point operations. Hence both the spatial and computational complexities are reduced notably by quantization. 

Binary neural network (BNN) is a typical aggressive quantization method \cite{courbariaux2016binarized}. The model weights and activations are expressed as $\{-1, +1\}$ that could be stored in only 1-bit. Benefiting from the hardware bitwise operations, the dot-product between weights and activations is replaced by XNOR and POPCOUNT arithmetics. Hence the deployment of BNN is no longer constraint by the GPUs. However, the naive BNN suffers from non-negligible performance degradation, especially on large scale and complicated tasks \cite{hubara2017qnn}. Although some proposed techniques have alleviated the information loss through improved binarization scheme, network topology and training algorithm, there still exists nontrivial accuracy gap between BNN and the full-precision network \cite{rastegari2016xnor} \cite{bethge2018training} \cite{hou2016loss}. Contemporarily, an effective method to boost the compact model performance is representing the model variables with fixed-point values, i.e. quantized neural network (QNN) \cite{hubara2017qnn}. As represented in \cite{jacob2018quantization-integer} \cite{choi2018pact}, the QNNs are able to achieve comparable accuracy as the full-precision networks under the circumstance of 4-bit quantization. Nevertheless, larger bitwidth means the linear increase of model size and higher requirement on hardware capacity. When the computing resources are extremely limited, it is necessary to make a trade-off between model accuracy and compression.

In this paper, we work on this trade-off issue by referring to mixed-precision approach. In fact, the network layers contribute diversely to the overall performance and each has different sensitivity to quantization. While the network propagating forward, the dissimilarity of hierarchical features is enhanced progressively. In the shallower layers, the internal features are distributed on complex manifolds. Accurate neurons are necessary to obtain the subsequent features. While in deeper layers, a rough convolutional filter is able to distinguish the previous local features as the deep semantic features are more separable. Hence the parameter precision could be designed flexibly based on the network structure and the distribution of hierarchical features. In this paper, a simple but effective QNN with progressively decreasing bitwidth is proposed. The original information is preserved well by the high-precision bottom layers while the model size is compressed further due to low-precision representation of top layers.

Our main contributions are:
\begin{enumerate}
	\item Based on the observation on internal feature distributions and network structure, a mixed-precision QNN framework with progressively decreasing bitwidths is proposed.
	\item Four typical classification CNNs including VGG, AlexNet and ResNet-18/20 and an object detection framework SSD are quantized based on the proposed mixed-precision method. The layer-wise bitwidth gradually reduces to 1-bit from 4-bit and 8-bit respectively.
	\item The redesigned QNNs are validated on several benchmark datasets including CIFAR-10/100, ILSVRC-2012 and PASCAL VOC. The experimental results demonstrate that the mixed-precision networks could achieve preferable or very similar performance while requiring at least 30\% less memory space for quantized parameters. 
\end{enumerate} 

The rest of this paper is organized as follows. Section 2 provides a summary of related works. Based on the analysis on the feature distribution of different layers, a multi-level quantized structure with gradually decreasing bitwidth is proposed in Section 3. In Section 4, we demonstrate the effectiveness of the mixed precision framework via  extensive experiments on several typical CNNs architectures and benchmark datasets. Section 5 ends this paper with some conclusions.

\section{Related Work}

Network compression and acceleration is critical to the practical deployment of CNNs on edge devices. One kind of paradigm focuses on the network topology structure. Some researches focus on the design of compact neural architecture. Many lightweight networks are proposed including ResNet \cite{he2016ResNet}, DenseNet \cite{huang2017densely}, MobileNet \cite{howard2017mobilenet} and ShuffleNet \cite{zhang2018shufflenet}. In addition, there exist some methods that search for an effective neural architecture via reinforcement learning \cite{zoph2018nasnet} \cite{tan2019efficientnet}. Some other researches conduct model compression from the opposite direction. A tiny network is obtained via pruning and sparsity constraints on the basis of a well-trained complex network \cite{molchanov2016pruning} \cite{frankle2018lottery} \cite{yu2018nisp}.

%网络量化则从数据结构的角度来进行。研究表明，在网络训练时，全精度的数据是没有必要的。通过将模型参数采用更低的精度进行表示，模型大小可以成倍降低。不仅如此，中间层特征同样可以被离散化，这样网络中的MAC就转为了定点运算。对于硬件的要求也大大降低，可以在特制平台上进行。因此受到了很多人的关注，也是本文的焦点。

\begin{figure}
\centering
\includegraphics[width=1.5in, keepaspectratio]{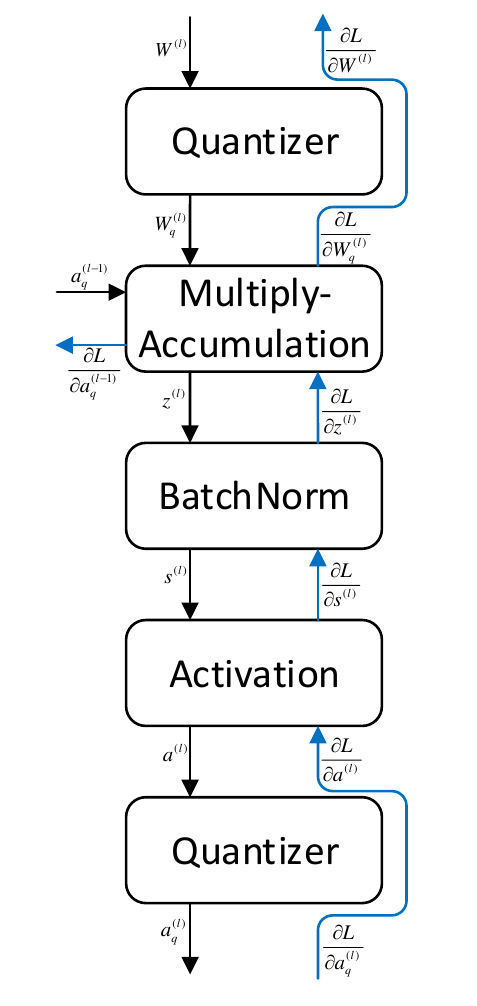}
\caption{The computation graph of neural cell in QNN. The black arrows depict the forward data flow and the blue ones show the backward-propagation. Both the full-precision and quantized values are remained during training. The un-differentiable quantizer module is bypassed in the computation graph. After training, full-precision weights are discarded in deployment.}
\label{fig: neural cell}
\end{figure}

Network quantization conducts the compression and acceleration task from the perspective of data format while preserving the network architecture. In \cite{gupta2015limited_precision}, the results shown that half-precision is also able to acquire promising accuracy. This indicated that the model parameters could be stored by lower bitwidth and the model size is reduced several times. More over, the intermediate variables are also could be represented by discrete values. Then the computationally expensive float-point arithmetics are replaced by fixed-point and bitwise operations which are able be conducted on dedicated hardware. As shown in Fig.~\ref{fig: neural cell}, both the full-precision and low-bit variables are preserved in the computation graph during the training phase. To make backward-propagation feasible, the gradients flow through the un-differentiable quantizer straightly, i.e. \textbf{s}traight-\textbf{s}hrough gradient \textbf{e}stimator (STE) \cite{bengio2013STE} \cite{hubara2017qnn}. Some training characters and theoretical analysis are demonstrated in \cite{tang2017train-compact} \cite{bethge2018train-scratch} \cite{alizadeh2019empirical} \cite{yin2019understandingSTE}. After training, it is unnecessary to reserve the full-precision values.

%二值化网络是一种aggressive的量化方式。模型参数和变量全部采用正负一来表示。这样只需要一个比特就是存放，将模型大小降低至1/32.其次所有计算只需XNOR、POPCOUNT操作即可完成。结果显示二值网络在简易任务上可以取得不错的效果，但是在复杂任务上精度损失很大。一些方法提出了更好的量化方式，但是与全精度网络的gap仍然不可忽视。Bi-real通过改进结构来提高性能，最终效果有限，并引入了大量的浮点数加法运算。

BNN is an aggressive form of network quantization. The weights and activations are expressed as $\{-1, +1\}$ according to the signs. Thus the memory space required for each variable is reduce to only 1-bit and the model size after binarization is nearly $1/32$ of the origin \cite{courbariaux2016binarized}. In addition, the inference efficiency is improved substantially by leveraging the XNOR and POPCOUNT operations \cite{hubara2017qnn}. However, the extreme compression also leads to heavy information loss during binarization. There exists non-trivial accuracy gap between BNN and the full-precision counterpart, especially on complicated tasks. Some techniques are proposed to alleviate the performance loss via modified binarization scheme \cite{rastegari2016xnor} \cite{hou2016loss} and network architecture \cite{liu2018bi-real}. These improvements are limited and extra full-precision arithmetic is introduced.

Another effective way to improve the model capability is assigning larger bitwidth to the network variables, i.e. conservative quantization \cite{hubara2017qnn}. A general and flexible quantization method is proposed in \cite{zhou2016dorefa} and achieves promising accuracy on ILSVRC-2012. \cite{choi2018pact} and \cite{esser2019learned} improve the QNN performance further by adjusting the quantization step size during back-propagation. In case of 4-bit quantization, the QNN could achieve comparable results as the full-precision counterpart. However, the increase of bitwidth means amplification of model size. There is a trade-off between superior performance and aggressive compression.

Among the methods mentioned above, all the model weights are treated equally and assigned with the same bitwidth. Actually, the parameters in the stacked neural network contribute differently to the overall results. It indicates that the parameter bitwidth should be determined by its individual function. More over, some advance chips are released including Apple A12 Bionic and Nvidia Turing GPU that support mixed-precision arithmetic. Hence some researches tackle the QNN trade-off issue via mixed-precision method. In \cite{fromm2018heterogeneous} and  \cite{zhou2018adaptive}, the bitwidth of each parameter is set according to the quantization residual of a pre-trained network. Wang \emph{et al.} \cite{wang2019haq} fine-tune the bitwidth via reinforcement learning.
In this paper, we explore the layer-wise bitwidth from another perspective and propose a simple but effective mixed-precision framework. In comparison with the previous work, this proposed method is more flexible and compatible with various quantization schemes. 
%The experiments validate that our method could obtain comparable results as $k$-bit homogeneous QNN while needing 30\% less bitwidth averagely for the quantized parameters.

\section{Methodology}

\subsection{Quantization Function}

As Fig.~\ref{fig: neural cell} shows, the discrete data flow through the stacked neural cells which consist of quantization, multiply-accumulation (MAC), batch normalization and activation. While the storage and computation cost is reduced notably, the information loss is inevitable due to quantization error during this process. An appropriate quantization module which is able to preserve the valuable information in the continuous variables is crucial for the network performance.
%
%In QNN, the discrete data flow through the stacked neural cells which consist of batch normalization, quantization function, and multiply-accumulation (MAC). Among them, MAC accounts for the majority of the total computation. By leveraging fixed-point arithmetics, the computational cost of MAC is reduced substantially. The deployment of QNN could be extended to the low-power hardwares other than GPUs. Hence an appropriate quantization function is required to transform the continuous variables to quantized values during training process.

\subsubsection{\textbf{Binarization}}
%In general, both the binarization and quantization functions are non-differentiable. This challenges the training of QNNs. To address this issue, the straight-through estimator (STE) is applied to approximate the gradient calculation \cite{courbariaux2016binarized}. Specifically, the quantization function is regarded as an identity mapping during back-propagation.

An extreme quantization method is to store the discrete value in 1-bit, i.e. binarization. Given a variable $x\in\mathbf{R}^{n}$, the binary value $x_{b}$ is determined by the sign. In order to improve the value range, a scaling factor $\frac{\lVert x \rVert_{1}}{n}$ is introduced. Then MAC is conducted by XNOR and POPCOUNT operations. However, the binarization function $B(\cdot)$ maps a continuous set $\mathbf{R}^{n}$ onto a discrete set $\{-1,+1\}^{n}$. The un-differentiability is an obstacle in the backward propagation and challenging the training of QNN. To address this issue, the STE is proposed to bypassing the quantizer \cite{bengio2013STE} \cite{hubara2017qnn}. 
The forward and backward computation of binarization is shown as the follows. $\mathbb{I}\{\cdot\}$ is the indicator function. If the condition satisfied, the indicator returns $1$. Otherwise, it returns 0.
%\begin{equation*}
%x_{b} = B(x) = \frac{\Vert x \Vert_{1}}{n}\text{sign}(x)
%\end{equation*}
\begin{align*}
\textbf{Forward: }&x_{b} = B(x) = \frac{\Vert x \Vert_{1}}{n}\text{sign}(x),\\
\textbf{Backward: }&\frac{\partial B}{\partial x} \approx \mathbb{I}\{|x|<1\}.
\end{align*}

\subsubsection{\textbf{Quantization}}
% 保守量化，即采用更大的比特，能够显著提高表示能力。一种常用的量化函数为
The conservative quantization can improve the model capacity significantly by utilizing larger bitwidth $k>1$. A general linear function $Q(\cdot)$ is defined as 
%\begin{equation*}
%x_{q} = Q(x) = \frac{1}{2^{k} - 1} \lfloor ( 2^{k} - 1 ) x \rceil,
%\end{equation*}
\begin{align*}
\textbf{Forward: } & x_{q} = Q(x) = \frac{1}{2^{k} - 1} \lfloor ( 2^{k} - 1 ) x \rceil, \\
\textbf{Backward: } & \frac{\partial Q}{\partial x} \approx 1,
\end{align*}
where $x\in [0, 1]^{n}$ and $x_{q} \in [0, 1]^{n}$ denote the full-precision and quantized values and $\lfloor \cdot \rceil$ represents the rounding operation. The STE gradient is utilized either in the backward of $Q(\cdot)$.
With this function, the model weights and activations could be discretized after proper preprocessing as follows.

\begin{figure}
	\centering
	\includegraphics[width=3.5in, keepaspectratio]{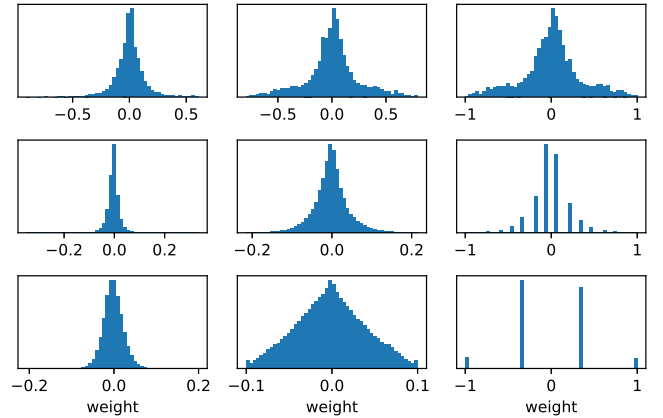}
	\caption{The the histogram of network weight parameters. The first column depicts the distribution of weights from three different layers in well-trained full-precision network. The second and third columns are the full-precision and quantized weights from the according layers in a well-trained QNN. The images from top to bottom in the third column represent the 8-bit, 4-bit and 2-bit quantization results respectively.}
	\label{fig: weight hist}
\end{figure}

\subsubsection{\textbf{Weight Quantization}} 
For a continuous weight tensor $W \in \mathbf{R}^{m\times n}$, it is necessary to project the unbounded elements into specified interval $[0, 1]$. The most straightforward normalization is scaling and shifting after dividing the largest absolute value. However, the majority of the continuous weight values distribute around the zero-point as Fig.~\ref{fig: weight hist} shows. The straightforward division would make the normalization dominant by the outliers and lead to additional round-off quantization error. Hence a non-linear transformation, hyperbolic tangent function, is introduced to alleviate the impact of long-tail distribution. Meanwhile, the saturation effect of $\tanh(\cdot)$ can suppress the variation of large values and avoid outliers during training. It is also worth noticing that the MAC operations are conducted channel-wise,
\begin{equation*}
z_{i}=W_{i} \cdot a, \quad W_{i}^{T}, a\in \mathbf{R}^{n}.
\end{equation*}
The MAC results are related to the weight values in the according channels. Hence it is more suitable to do channel-wise normalization. The extra scaling factors can be merged into the batch normalization parameters and no addition computation cost is introduced during deployment. Thus the overall quantization process for weights is as follows,
\begin{gather*}
\hat{W} = \tanh(W) ,\\
M_{i} = \max_{j}(|\hat{W}_{ij}|) ,\\
W_{q,ij} = 2\cdot Q( \frac{\hat{W}_{ij}}{2\cdot M_{i}} + \frac{1}{2} ) - 1.
\end{gather*}

\subsubsection{\textbf{Activation Quantization}}
%对于激活值，同样可以采取类似的正规化方法。但和weight不同的是，这样的操作在推断时会带来大量的额外运算，降低模型效率。
For the activation quantization, it is theoretically feasible to adopt the similar strategy as weight parameters. But the model efficiency will be reduce badly due the additional float-point operations in preprocessing. Therefore a clamp function is usually applied as activation function to confine the features to specified interval $[0, 1]$ before quantization.
%\begin{equation*}
%a_{q} = Q( \text{clip}(a, 0, 1) ).
%\end{equation*}
\begin{gather*}
a = \text{clamp}(s, 0, 1),\\
a_{q} = Q(a).
\end{gather*}

\subsection{Hierarchical Feature Distribution and Mix Precision QNN}
\label{Sec: 3.2}

\begin{table*}[t]
	\centering
	\caption{number of weight parameters in typical networks}
	\begin{tabular}{cccccccc}
		\toprule
		Layer & 1     & 2     & 3     & 4     & 5     & 6     & 7 \\
		\midrule
		VGG-7 & 3,456  & 147,456 & 294,912 & 589,824 & 1,179,648 & 2,359,296 & 8,388,608 \\
		ResNet-20 & 432   & 13,824 & 51,200 & 204,800 & -     & -     & - \\
		AlexNet & 41,472 & 307,200 & 884,736 & 663,552 & 442,368 & 37,748,736 & 16,777,216 \\
		ResNet-18 & 1,728  & 147,456 & 524,288 & 2,097,152 & 8,388,608 & -     & - \\
		\bottomrule
	\end{tabular}%
	\label{tab:num_para}%
\end{table*}%
%1. 神经网络可以自动地学习特征表示。提取到的特征越来越好，即不用类别样本的特征分布逐渐分化，相同类别的特征分布逐渐聚拢。最终特征从高维流形分布降低至一个低维线性分布。在这一过程中，每个神经元都相当于一个简易分类器。
One of the important advantaged that contributes to the remarkable achievements of deep neural network is that a delicate feature representation could be learned automatically by end-to-end training. Based on the network topology structure, the hierarchical features are improved by MAC operation and non-linear transformations layer-wise. As the network propagates forward, the variation of each categorical feature distribution is reduced gradually while the margins between each other increase. Consequently, the feature distributions are mapped from complex manifolds in high-dimension to several clusters in low-dimension and a linear classifier is able to achieve great accuracy by leveraging the final semantic features.

%One of the important advantages of deep neural network that contribute to the remarkable achievements is that it could learn a delicate feature representation automatically. Based on the network topology, the hierarchical features are obtained by the inner-product operations and non-linear activations layer-wise. While the network propagating forward, the linear separability of feature distribution is enhanced. Consequently a linear classifier is able to achieve great accuracy by leveraging the final semantic features. During this process, each neuron works as a basic classifier to extract target feature. The feature distributions of different categories is mapped from complex manifolds to several clusters in high dimension space.
%As the linear separability of features is improved, we argue that the requirement on neuron capability is reduced. 

%2. 实际特征分布展示

\begin{figure}[h]
	\centering 
	\subfigure[]{\label{fig1:subfig:a}
		\includegraphics[width=0.45\linewidth]{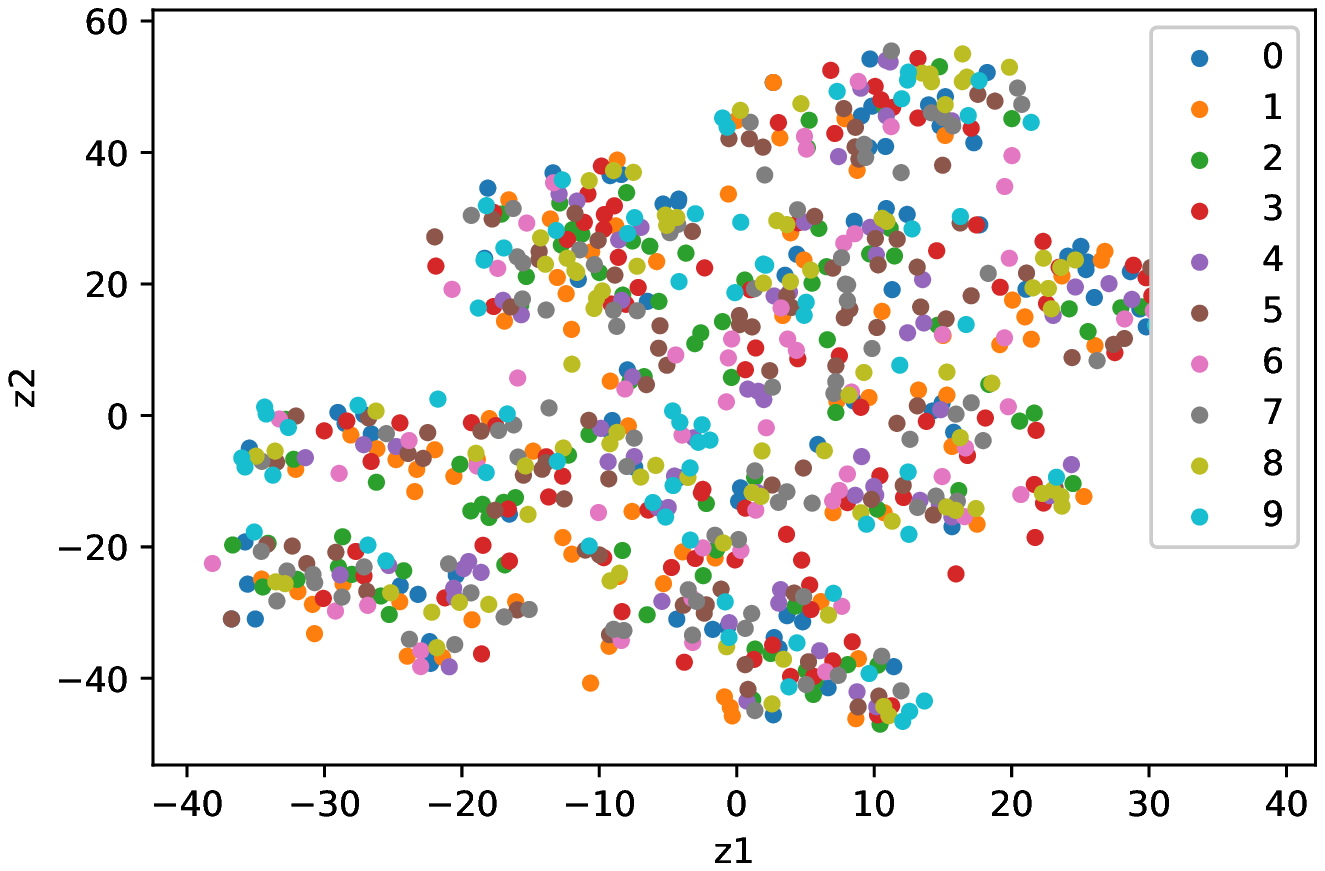}}
	\hspace{0.01\linewidth}
	\subfigure[]{\label{fig1:subfig:b}
		\includegraphics[width=0.45\linewidth]{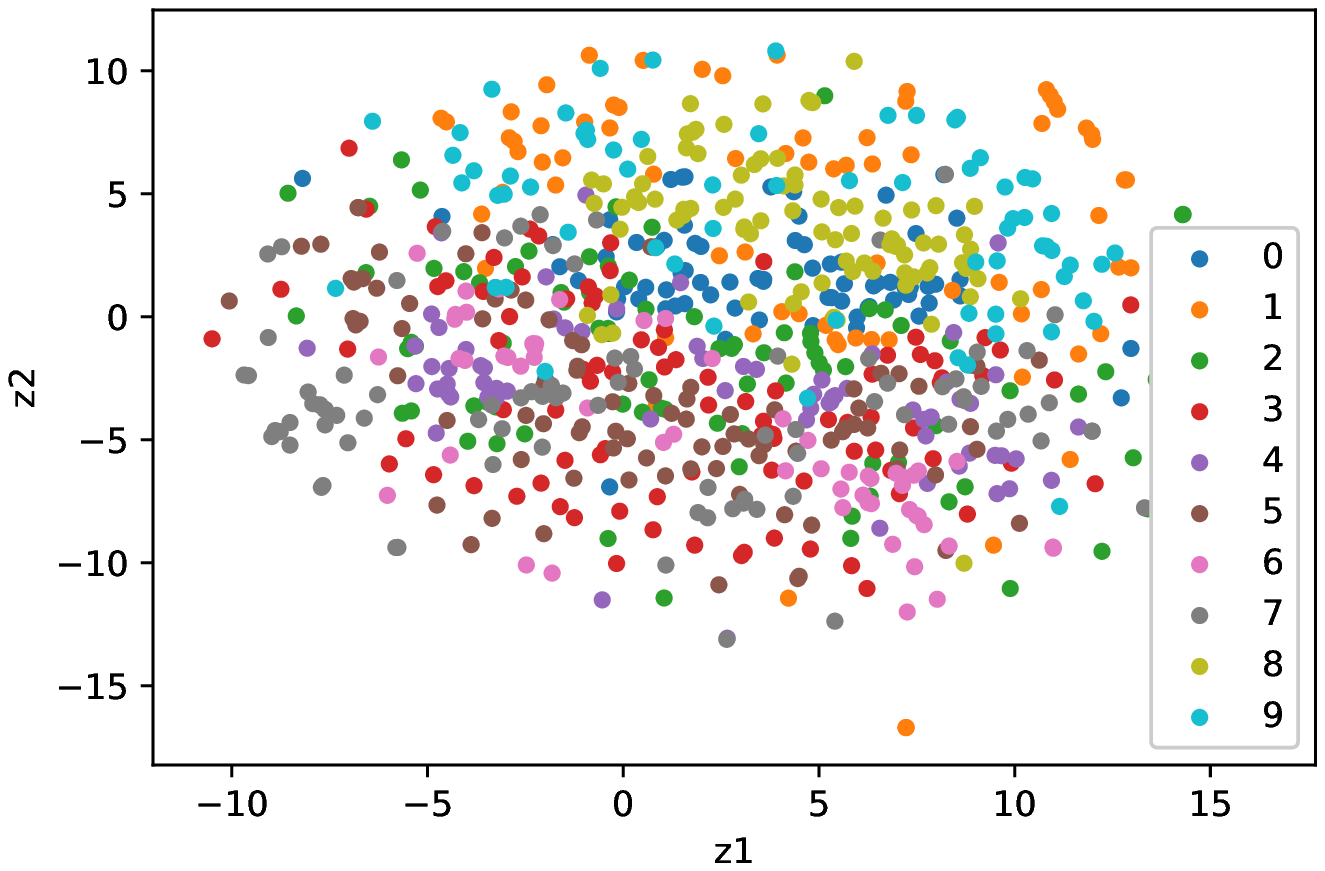}}
	\vfill
	\subfigure[]{\label{fig1:subfig:c}
		\includegraphics[width=0.45\linewidth]{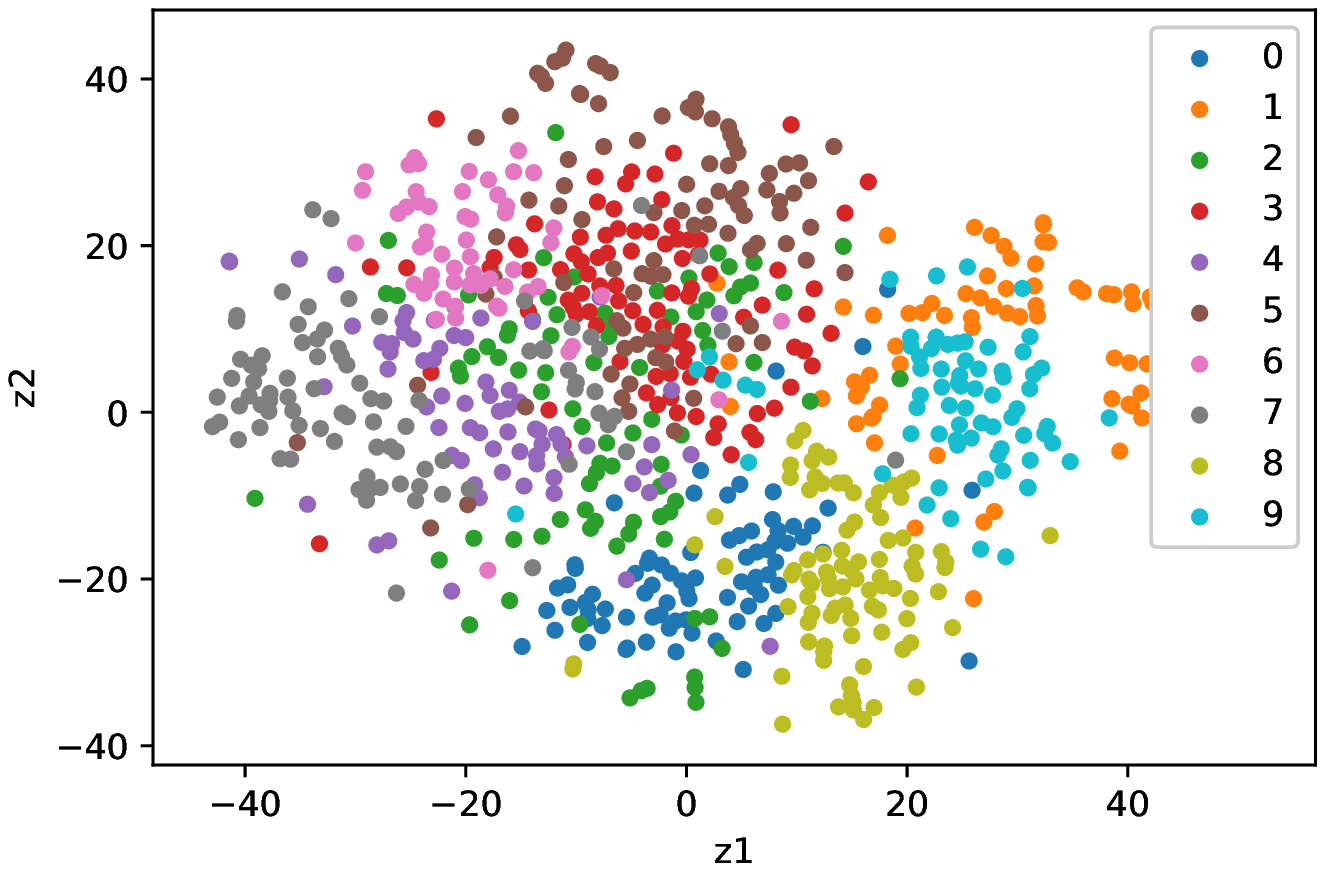}}
	\hspace{0.01\linewidth}
	\subfigure[]{\label{fig1:subfig:d}
		\includegraphics[width=0.45\linewidth]{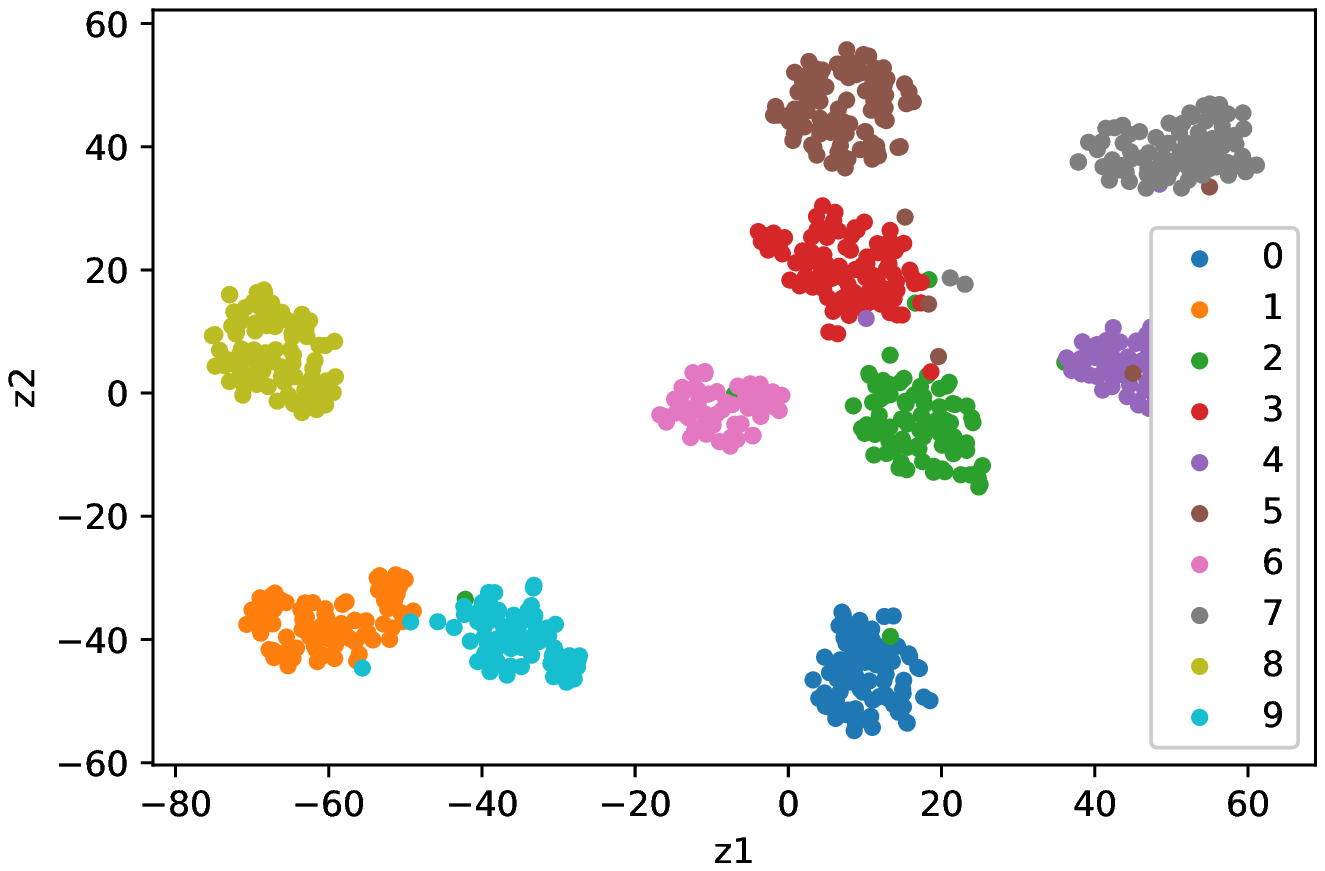}}
	\caption{In the bottom layer of a trained network, the feature distributions of different categories intervenes each other severely as (a) shows. Many delicate neurons are needed to distinguish the overlapped distributions. And as the network propagates forward, the feature distribution of the same category gathers gradually in (b) and (c). In the end of the hidden layers, there exist clear margins between the semantic feature distributions of different classes in (d). 
		%For more separable feature distributions, a simple neuron with lower-precision parameters is able to extract meaningful semantic features.
		With the improvement of separability among the feature manifolds, a neuron with lower-precision parameters is able to extract robust feature.
	}
	\label{fig1:feature distribution}	
\end{figure}

To illustrate this intuition explicitly, a VGG-7 network\footnote{VGG-7 architecture: 2$\times$(128-Conv3$\times$3) + MP2 + 2$\times$(256-Conv3$\times$3) + MP2 + 2$\times$(512-Conv3$\times$3) + MP2 + 1024-FC + Output-FC \label{VGG}} model which consists of 6 convolutional layers and 1 latent fully-connected layers is trained based on the CIFAR-10 dataset. After training, 50 samples from each class are selected randomly and fed into the model. The feature representations in the internal layers are extracted and transformed into 2-dimension by t-SNE \cite{maaten2008t-SNE}.
As shown in Fig.~\ref{fig1:subfig:a}, there exists severe aliasing among the feature distributions of different categories after dimension reduction in the initial layer. It indicates that the feature manifolds in the bottom layer is very complicated. This is due to that the elementary characteristics of images are mail color and texture features. Theses local representations are quite similar with each other of different samples. It is difficult to determine the labels based on the elementary features directly. Many delicate neurons is required to distinguish the overlap distributions. As the network propagates forward, the features of the same category become organized gradually. 
As Fig.~\ref{fig1:subfig:d}shows, the advanced semantic features are more robust and there exist clear margins between the distribution clusters in deep layers. 
%It means that a coarse neuron is still able to extract the subsequent advanced feature.

%3. 由于特征分布越来越简单，这意味着对每一层神经元的精度要求是不一样的。浅层必须从及其复杂的分布中提取特征，因此要求性能更高，对量化更加敏感。当浅层提取到足够好的特征后，后续的神经元性能要求逐渐降低，可以采用更低的比特。因此，网络的结构可以更灵活的设计。同时我们关注到，模型的参数主要集中在深层。这样的结构相对于低比特网络相比，并不会增加显著模型大小，但是有潜力获取更好的浅层特征，为网络整体的性能奠定基础。而同高比特网络相比，减少深层比特，可以显著降低模型大小。

%During this process, each neuron works as a simple classifier to extract target feature.
During this process, the feature transformation is conducted by neurons in each layer. Every neuron works as a simple classifier to extract target feature. The input complexities of the network layers differ with each other, which means that the precision requirement on the neurons are also different. Based on this observation, we argue that the neurons in the shallower layers are more sensitive to quantization. As the feature distributions overlap mutually, finite neuron are unable to distinguish the samples and extract meaningful intermediate representations without suitable precision. Once the advanced features are obtained explicitly, the following layers become more robust to the quantization error. Thus it is feasible to design the QNN structure more flexibly rather than $k-$bit homogeneous networks. The bitwidth for each model parameter progressively decreases from the initial $k$-bit as the QNN propagates forward.

It also worth noticing that the majority of model parameters are concentrated in the deeper layers as Table~\ref{tab:num_para} shows. The rise of bitwidth at the bottom layers has little effect on the model size in comparison with low-bit network. But the original information would be preserved better. On the other hand, the model size of mixed-precision QNN is much smaller than the $k-$bit homogeneous ones due to lower parameter precision. Hence the mixed-precision QNN is more compact and has the potential to achieve promising performance.

By utilizing the framework of progressively decreasing bitwidth, 4 typical CNNs are quantized. VGG-Net and AlexNet are the representatives of plain CNNs. The VGG-7 in this paper is designed for CIFAR-10/100 dataset. All the weight parameters are quantized except that of the output layer as the linear classifier is related to the final results directly and requires enough precision. The bitwidth for the quantized layers decreases from 8-bit to 1-bit layer-wise with a factor $1/2$ as shown in Table~\ref{tab:cifar10}. Although the initial bitwidth is higher than the homogeneous counterpart, the average model bitwidth is reduced to 1.06. AlexNet, which contains 5 convolution and 2 latent fully-connected layers, is proposed for the high-resolution image recognition task ILSVRC-2012 \cite{krizhevsky2012alexnet}. The input and output layers are maintained full-precision as \cite{rastegari2016xnor} \cite{zhou2016dorefa} for a fair comparison. The bitwidth setting is similar with VGG-7 and shown in Table~\ref{tab:imagenet}. The final average bitwidth is 1.10.

ResNet is the pioneer of networks with shortcuts. The ResNet-20, which consists of 3 residual stages, is initially proposed for the CIFAR-10 task \cite{he2016ResNet}. For a fair comparison with related work \cite{rastegari2016xnor} \cite{zhou2016dorefa}, the convolutional weights of residual stages are quantized from 4-bit to 1-bit as shown in Table \ref{tab:cifar10}. 
As ResNet-20 has only 64 filters at the final stage, it uncertain that the 64-dim pooling features obtained by aggressively quantized neurons could satisfy the classification requirement, especially for CIFAR-100 task. A doubled bitwidth model with more powerful capacity is also validated in this paper. 
By contrast, The ResNet-18, which contains 4 residual stages, is much wider and has 512 filters at the final residual stage. The bitwidth reduces from 8-bit to 1-bit as shown in Table \ref{tab:imagenet}.
The activation bitwidth of the mixed-precision network is set the same with the homogeneous counterparts.

The object detection is much more complicated task than image classification. In addition to predict categories of multiple object in an image, the network also needs to regress coordinates of the bounding boxes. This requires higher feature extract capability of the network. To investigate the performance of mixed-precision QNN on object detection task, a VGG-16 based single shot detector (SSD) \cite{liu2016ssd:} is quantized in this paper. The weight parameters of VGG-16 backbone is discretized utilizing the similar bitwidth setting as VGG-7. To improve the feature extraction capability at the final stage, the bitwidth of extra layers is set to 4-bit. The output layers remained full-precision. The final average bitwidth is 1.42.

\section{Experiments}

To validate the performance of QNN with progressively decreasing bitwidth, we conduct extensive experiments on CIFAR-10/100, ILSVRC-2012 and Pascal VOC datasets. 
%The experiments are performed on a work station with an Intel Core i7-7800X CPU, 32 GB RAM and 2$\times$GeForce GTX 1080Ti GPUs.

\subsection{CIFAR-10/100}
There are 10 classes of 50,000 training images and 10,000 test ones in CIFAR-10 dataset. The image size is $32 \times 32$ pixels. The CIFAR-100 dataset consists of the same number of images from 100 categories. One tenth of training samples is selected as validation set. 

%A VGG-Net and a ResNet-20 \cite{he2016ResNet} are defined to validate the performance of the proposed method. As the ConvNet is a quite wide network, these exist sufficient feature representations in each layer. All the weight parameters are quantized and except the last layer due to that the linear classifier obtains the final results directly and must has enough precision. On the contrary, ResNet-20 is much thiner and has only 16 filters in the first residual block. In oder to preserve the image information, the first and last layers both remain full-precision as the previous work \cite{zhou2016dorefa}. All the activation in the two networks are stored with the same bitwidth. The detailed bitwidth setting for each layer is shown in Table \ref{tab:cifar10} and Table \ref{tab:cifar100}. 
%FP and 32-bit represent the full-precision networks with float-point parameters. And the mean bitwidth $k_{w}$ for weights is calculated through dividing the total bit space of quantized weights 

We follow the data augmentation in \cite{he2016ResNet} for training. At testing time, the original images are sampled directly. We use SGD optimizer with momentum of 0.9 and learning rate starting from 0.1 and scaled by 0.1 at epoch 80, 120, 160. L2-regularizer with decay of 2e-4 is applied to weight parameters. The mini-batch size is 128 and after 200 epochs of training from scratch, the test accuracy associated with the best validation performance is reported as the final result.

\begin{figure}[h]
	\centering
	\includegraphics[width=3.5in]{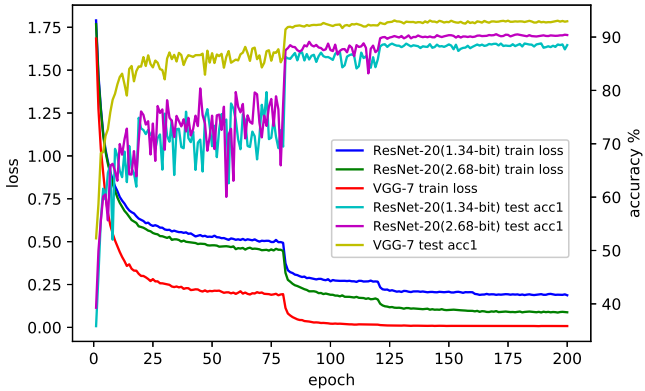}
	\caption{The training curve of ResNet-20 and VGG-7 on CIFAR-10.}
	\label{fig: cifar10}
\end{figure}

After 5 runs of each experiment, the average test accuracies of CIFAR-10 are recorded in Table \ref{tab:cifar10}. Here, FP and 32-bit denote the full-precision network with float-point parameters.
As the analysis in Sec.~\ref{Sec: 3.2}, the mixed-precision networks obtain higher accuracies than the homogeneous counterparts while the model size is smaller. For ResNet-20, the re-designed network with less than 3-bit for weights and 4-bit for activations is able to achieve comparable final result as the full-precision network. However, at beginning the training process, the generalization ability of mixed-precision QNN fluctuates obviously as Fig.~\ref{fig: cifar10} shows. This is due to that the quantized values change back and forth due to large learning rate. When the learning rate decays, the training process become stable.
In addition, the mixed-precision VGG-Net obtains better result than both the 2-bit and even the full-precision one. 
We argue that the better information preservation in the initial layers due to higher bitwidth boosts the performance evidently. Meanwhile, the VGG-7 is a very ``wide'' network. The redundancy stabilizes the training process as Fig.~\ref{fig: cifar10} shows. But once sufficient and meaningful information is obtained by the bottom layer, the redundant parameters in the subsequent layers may lead to overfitting. Hence, the suitable bitwidth setting contribute to the model regularization.

%As the analysis before, the multiple bitwidth networks achieve the best performance in all four scenarios while requires less than 1.5 bit for each parameter. In addition, even if only the bitwidth of bottom layers is enhanced, the network could obtain comparable of better results in comparison with 2-bit homogeneous network. We argue that when the bitwidth is high in shallow layers, the image information is preserved adequately. In the deep layers, the lower-precision contribute to model regularization.

\begin{table}[h]
	\centering
	\caption{CIFAR-10 Experimental Results}
	\begin{tabular}{ccccc}
		\toprule
		Model & Method & $k_{w}$  & $k_{a}$  & Test Acc. \% \\
		\midrule
		\multirow{5}[6]{*}{ResNet-20} & FP \cite{he2016ResNet}   & 32    & 32    & 91.60 \\
		\cmidrule{2-5}          & \multirow{2}[2]{*}{DoReFa \cite{zhou2016dorefa}} 
		%		& 2     & 32    & 90.51 \\
		%		&       & 4     & 32    & 90.97 \\
		%		&       
		& 2     & 2     & 88.20 \\
		&       & 4     & 4     & 90.50 \\
		\cmidrule{2-5}          & \multirow{2}[2]{*}{Ours} 
		%		& 1.34 (4-2-1) & 32    & 90.54 \\
		%		&       & 2.68 (8-4-2) & 32    & 91.32 \\
		%		&       
		& 1.34 (4-2-1)  & 2     & 88.33 \\
		&       & 2.68 (8-4-2)  & 4     & 90.54 \\
		\midrule
		\multirow{6}[10]{*}{VGG-7} & FP    & 32    & 32    & 92.48 \\
		\cmidrule{2-5}          & BNN \cite{courbariaux2016binarized}  & 1     & 1     & 89.85 \\
		\cmidrule{2-5}          & HWGQ \cite{cai2017hwgq}  & 1     & 2     & 92.51 \\
		\cmidrule{2-5}          & \multirow{2}[2]{*}{DoReFa \cite{zhou2016dorefa}} & 1     & 2     & 92.33 \\
		&       & 2     & 2     & 92.83 \\
		\cmidrule{2-5}          & Ours  & 1.06 (8-4-2-1-1-1/1)  & 2     & 93.22 \\
		\bottomrule
	\end{tabular}%
	\label{tab:cifar10}%
\end{table}%

The results on CIFAR-100 dataset are recorded in Table \ref{tab:cifar100} and consistent with that of CIFAR-10 generally. It is noticeable that our ResNet-20 result at the forth line is 3\% lower than the homogeneous bitwidth network. The reason is that ResNet-20 is a very ``narrow'' network that originally designed for CIFAR-10. After the average pooling layer, the dimension of semantic feature, 64, is less than that number of classes. Hence the 1-bit neurons in deep layers would induce significant information loss. Once the overall bitwidth is increased, the performance bottleneck is broken. While for the wide network, VGG-Net, it is unnecessary to worry about this. The numerous 1-bit neurons in deep layer guarantee meaningful semantic features. In comparison with the 2-bit network, the mixed-precision model is able to compress memory space for quantized parameters to nearly a half while achieving very competitive accuracy.

\begin{table}[h]
	\centering
	\caption{CIFAR-100 Experimental Results}
	\begin{tabular}{ccccc}
		\toprule
		Model & Method & $k_{w}$  & $k_{a}$  & Test Acc. \% \\
		\midrule
		\multirow{5}[6]{*}{ResNet-20} & FP    & 32    & 32    & 66.29 \\
		\cmidrule{2-5}          & \multirow{2}[2]{*}{DoReFa \cite{zhou2016dorefa}} 
		%		& 2     & 32    & 64.84 \\
		%		&       & 4     & 32    & 65.95 \\
		& 2     & 2     & 60.42 \\
		&       & 4     & 4     & 63.86 \\
		\cmidrule{2-5}          & \multirow{2}[2]{*}{Ours} 
		%		& 1.34 (4-2-1) & 32    & 63.61 \\
		%		&       & 2.68 (8-4-2) & 32    & 65.65 \\
		& 1.34 (4-2-1) & 2     & 57.82 \\
		&       & 2.68 (8-4-2) & 4     & 63.36 \\
		\midrule
		\multirow{5}[8]{*}{VGG-7} & FP    & 32    & 32    & 72.03 \\
		\cmidrule{2-5}          & XNOR \cite{rastegari2016xnor} & 1     & 1     & 57.74 \\
		\cmidrule{2-5}          & \multirow{2}[2]{*}{DoReFa \cite{zhou2016dorefa}} & 1     & 2     & 69.64 \\
		&       & 2     & 2     & 71.44 \\
		\cmidrule{2-5}          & Ours (8-4-2-1-1-1/1)  & 1.06  & 2     & 71.53 \\
		%		\midrule
		%		\multirow{5}[6]{*}{ResNet-18 \cite{he2016ResNet}} & FP    & 32    & 32    & 75.61 \\
		%		\cmidrule{2-5}          & \multirow{2}[2]{*}{DoReFa \cite{zhou2016dorefa}} & 2     & 2     & 71.09 \\
		%		&       & 4     & 4     & 73.31 \\
		%		\cmidrule{2-5}          & \multirow{2}[2]{*}{Ours} & 1.09 (4-2-1-1) & 2     & 72.43 \\
		%		&       & 1.42 (8-4-2-1) & 4     & 73.52 \\
		\bottomrule
	\end{tabular}%
	\label{tab:cifar100}%
\end{table}%

\begin{figure}[h]
	\centering
	\includegraphics[width=3.5in]{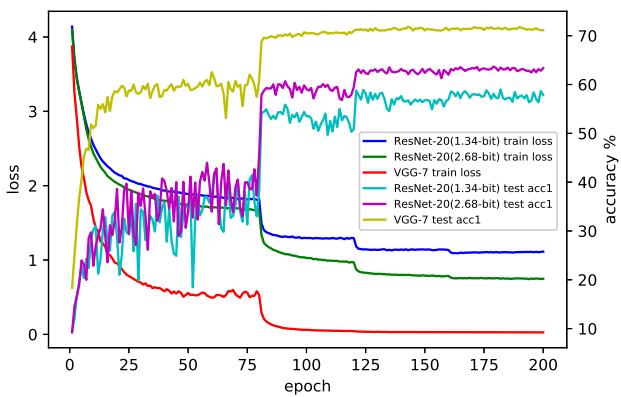}
	\caption{The training curve of ResNet-20 and VGG-7 on CIFAR-100.}
	\label{fig: cifar100}
\end{figure}

\subsection{ILSVRC-2012}

ILSVRC-2012 is a 1000-category dataset which consists of 1.2 million training images and 50 thousands of validation ones. Compared to CIFAR task, ILSVRC is much more challenging due to larger and diverse images. For training, the images are resized to $256 \times 256$ and cropped randomly to $224 \times 224$. For validation, the center crops are used as inputs.

%We adapted the proposed method to two typical networks on ILSVRC: AlexNet and ResNet-18. In order to make it compatible with QNN training, the original AlexNet is modified by removing the dropout layers and integration with batch normalization in our experiments. The quantized bitwidth of each layer is shown in Table \ref{tab:imagenet}. For a fair comparison, the first and last layers of the selected networks are not quantized as \cite{rastegari2016xnor} \cite{zhou2016dorefa}.
%In comparison with previous experiments, the initial bitwidth start from 8-bit due to more complex task. The activation setting is the same with CIFAR-10/100 experiments.

In the training process, an Adam optimizer with learning rate of 2e-4 and no weight-decay is applied for AlexNet. For ResNet-18, a SGD optimizer with learning rate of 0.1 and weight-decay of 1e-4. The learning rate is scaled by 0.1 at 60 and 75 of the 90 total epochs and at 30, 60, 90 and 100 of 120 total epochs respectively. After training, the Top-1 validation accuracies are reported in Table~\ref{tab:imagenet}. It is clearly that the mixed-precision QNNs have advantages over the ordinary ones in terms of both performance and model size. In comparison with the full-precision networks, the results are still acceptable.

\begin{table}[t]
	\centering
	\caption{ILSVRC-2012 Experimental Results}
	\begin{tabular}{ccccc}
		\toprule
		Model & Method & $k_{w}$  & $k_{a}$  & Top1 Acc. \% \\
		\midrule
		\multirow{4}[2]{*}{AlexNet} & FP    & 32    & 32    & 56.60 \\
		& XNOR \cite{rastegari2016xnor}  & 1     & 1     & 44.20 \\
		& DoReFa \cite{zhou2016dorefa} & 1     & 2     & 47.70 \\
		& Ours  & 1.10 (8-4-2-1/1-1)  & 2     & 53.18 \\
		\midrule
		\multirow{5}[2]{*}{ResNet-18} & FP    & 32    & 32    & 69.30 \\
		& XNOR \cite{rastegari2016xnor} & 1     & 1     & 51.20 \\
		& Bi-Real \cite{liu2018bi-real} & 1     & 1     & 56.40 \\
		& DoReFa \cite{zhou2016dorefa} & 2     & 2     & 62.60 \\
		& PACT \cite{choi2018pact}   &  2      & 2     & 64.40 \\
		& Ours  & 1.42 (8-4-2-1)  & 2     & 65.03 \\
		\bottomrule
	\end{tabular}%
	\label{tab:imagenet}%
\end{table}%

\begin{figure}[h]
	\centering
	\includegraphics[width=3.5in]{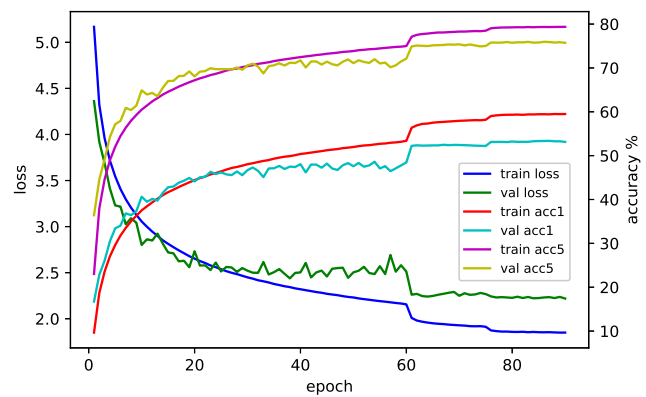}
	\caption{The training curve of AlexNet.}
	\label{fig: alexnet}
\end{figure}

\begin{figure}[b]
	\centering
	\includegraphics[width=3.5in]{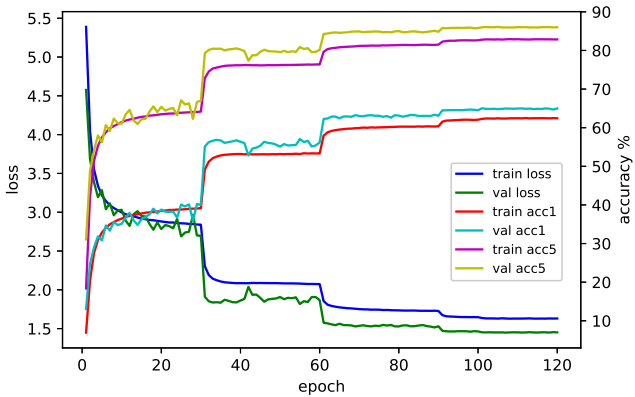}
	\caption{The training curve of ResNet-18.}
	\label{fig: resnet18}
\end{figure}

\subsection{Pascal VOC}

Pascal VOC is a benchmark dataset for object detection, which consist of 20 categories of objects in general. To validate the performance of the proposed method on more challenging tasks, we select SSD as a baseline detector and train our model on VOC2007 trainval and VOC2012 trainval datasets (16,551 images) after quantization. Then resulted model is evaluated on the VOC2007 test dataset (4,952 images). Our quantized models are trained from scratch without pre-training on ILSVRC dataset. An SGD optimizer with weight-decay of 1e-4 is applied for 8,000 iterations of training. The learning rate 1e-3 is used for the first 4,000 iterations, then continue training for 2,000 iterations with 1e-4 and 1e-5.

The comparison results are illustrated in Table~\ref{tab:pascal voc}. Compared with the full-precision counterpart, the performance of quantized networks degrade significantly due to more challenging task and quantization error. However, the mixed-precision network is still outperform the homogeneous one. In addition, the 62.21\% mAP means that the quantized detector has basic capabilities for object detection. From the detailed results and demo samples, we can conclude that the mixed-precision detector perform well on the object which are large enough and located at the center of images.
\begin{table*}[t]
	\scriptsize
	\centering
	\caption{Pascal VOC Experimental Results}
	\resizebox{\textwidth}{8mm}{
	\begin{tabular}{cccccccccccccccccccccccc}
%	\begin{tabular}{|c|c|c|c|c|c|c|c|c|c|c|c|c|c|c|c|c|c|c|c|c|c|c|c|}
		\toprule
		Model & Method & kw    & ka    & mAP   & areo  & bike  & bird  & boat  & bottle & bus   & car   & cat   & chair & cow   & table & dog   & horse & mbike & person & plant & sofa  & train & tv \\
		\midrule
		\multirow{3}[2]{*}{SSD-300} & FP    & 32    & 32    & 75.10 & 76.92 & 82.08 & 74.83 & 68.10 & 47.80 & 83.23 & 83.99 & 88.26 & 56.65 & 79.88 & 74.56 & 85.80 & 84.96 & 81.48 & 76.39 & 43.40 & 77.08 & 87.57 & 75.14 \\
		& Dorefa & 2     & 2     & 60.66 & 67.06 & 73.52 & 47.77 & 50.39 & 22.89 & 70.48 & 78.28 & 72.27 & 41.93 & 57.04 & 63.61 & 66.49 & 75.54 & 74.63 & 68.57 & 25.59 & 63.48 & 75.61 & 59.83 \\
		& Ours  & 1.22  & 2     & 62.21 & 70.96 & 76.08 & 51.64 & 54.97 & 25.33 & 72.37 & 78.79 & 74.07 & 44.30 & 56.27 & 62.43 & 66.91 & 78.71 & 73.82 & 69.57 & 27.19 & 64.00 & 77.20 & 60.69 \\
		\bottomrule
	\end{tabular}}%
	\label{tab:pascal voc}%
\end{table*}%

\begin{figure*}[t]
	\centering 
	\subfigure[]{\label{fig:subfig:b}
		\includegraphics[height=1.5in]{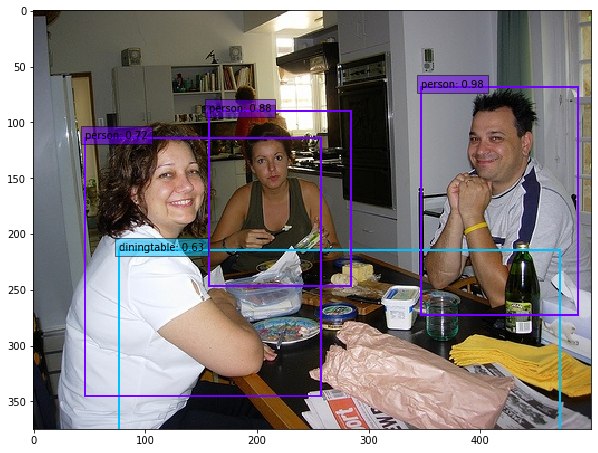}}
	\hspace{0.01\linewidth}
	\subfigure[]{\label{fig:subfig:c}
		\includegraphics[height=1.5in]{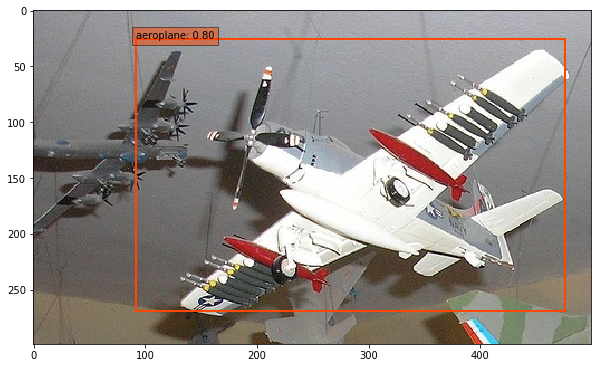}}
	\hspace{0.01\linewidth}
	\subfigure[]{\label{fig:subfig:d}
		\includegraphics[height=1.5in]{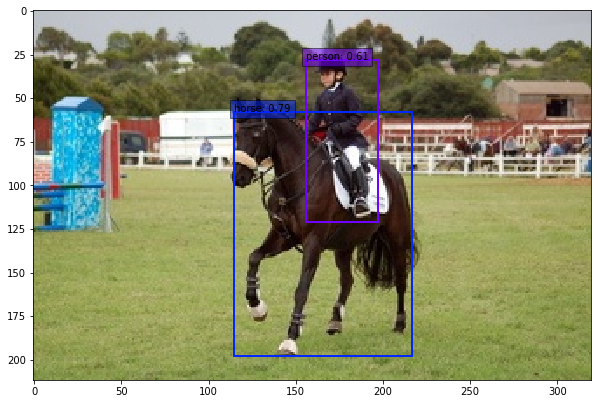}}
	\hspace{0.01\linewidth}
	\subfigure[]{\label{fig:subfig:e}
		\includegraphics[height=1.5in]{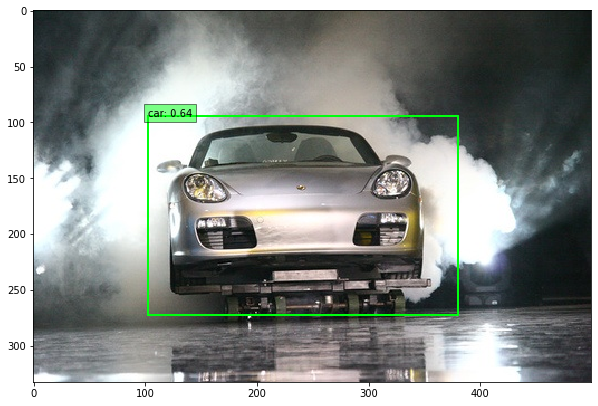}}
	\hspace{0.01\linewidth}
	\subfigure[]{\label{fig:subfig:e}
		\includegraphics[height=1.5in]{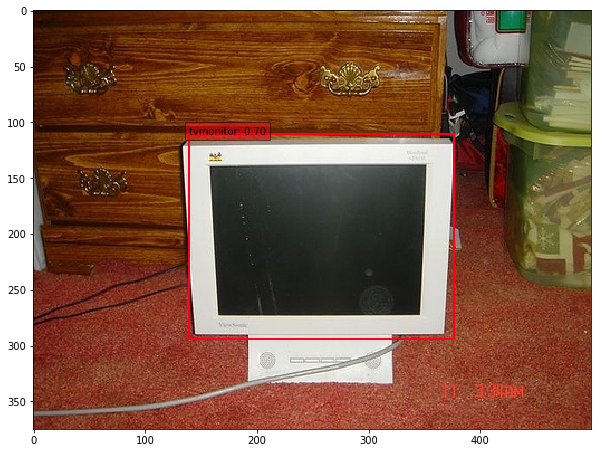}}
	\hspace{0.01\linewidth}
	\subfigure[]{\label{fig:subfig:e}
		\includegraphics[height=1.5in]{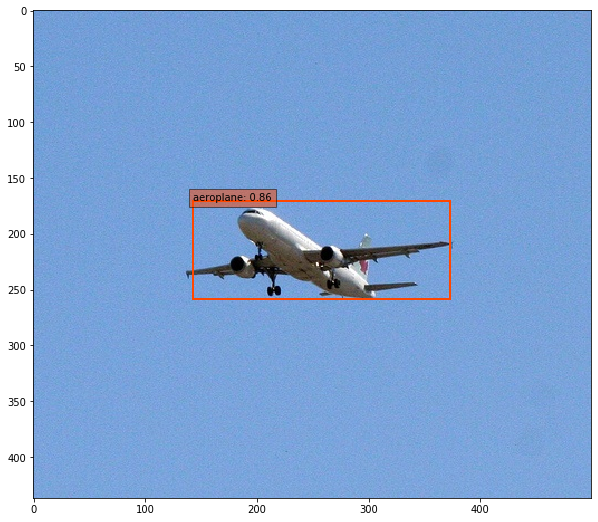}}
	\hspace{0.01\linewidth}
	\subfigure[]{\label{fig:subfig:e}
		\includegraphics[height=1.5in]{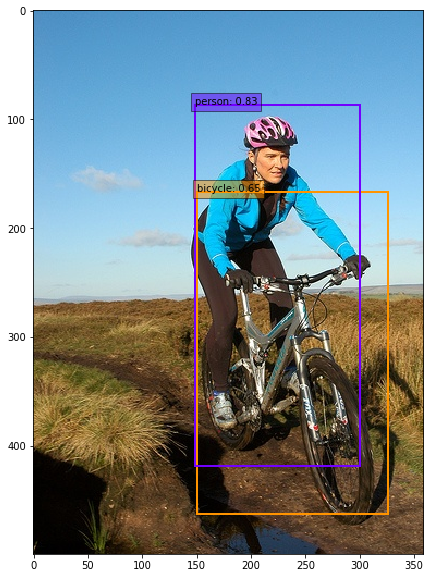}}
	\hspace{0.01\linewidth}
	\subfigure[]{\label{fig:subfig:e}
		\includegraphics[height=1.5in]{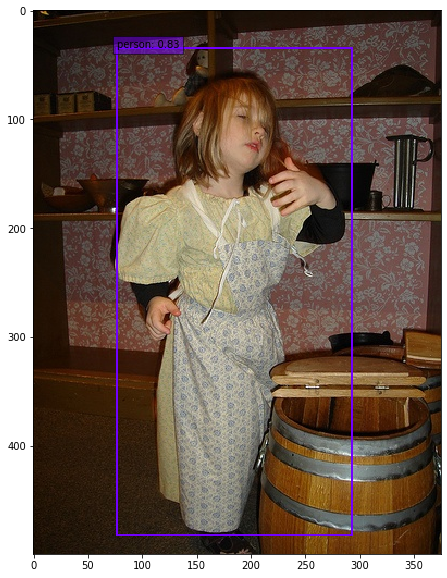}}
	\hspace{0.01\linewidth}
	\subfigure[]{\label{fig:subfig:e}
		\includegraphics[height=1.5in]{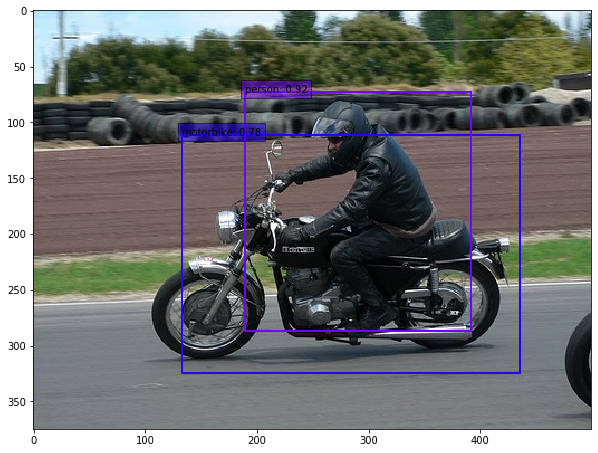}}
	\hspace{0.01\linewidth}
	\subfigure[]{\label{fig:subfig:e}
		\includegraphics[height=1.5in]{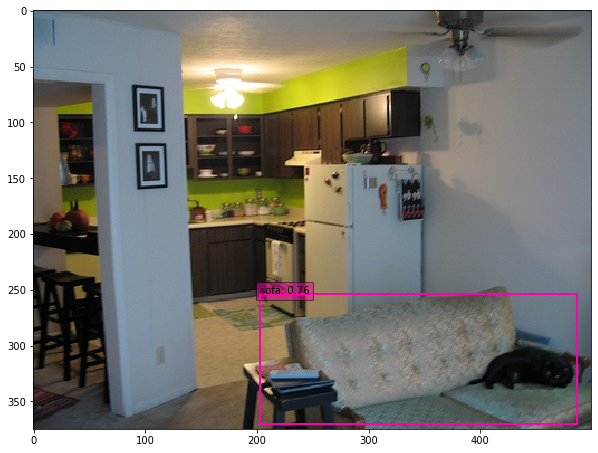}}

	\hspace{0.01\linewidth}
	%	\centering
	\caption{The sampled detection results of the mixed-precision SSD.	}
	\label{fig:detection samples}	
\end{figure*}

\section{Conclusions}

In this paper, a novel QNN framework with multiple bitwidth is proposed. Based on the observation of layer-wise feature distributions and network structure, we define a gradually decreasing bitwidth setting to preserve the original image information in bottom layers and address the trade-off between accuracy and compression. 
Extensive experiments on typical network architectures and benchmark datasets demonstrate that the proposed mixed-precision QNN could achieve preferable results in comparison with $k$-bit homogeneous networks while requiring 30\% less memory space for quantized parameters.

\section*{Acknowledgment}

This work is jointly supported by the National Natural Science Foundation of China (61603248) and the State Grid Corporation (Research and Demonstration Application of Monitoring and Management Technology of City Energy System Based on Large Data and Artificial Intelligence CEGHJS1800002).

\ifCLASSOPTIONcaptionsoff
  \newpage
\fi



\bibliographystyle{IEEEtran}
\bibliography{mybibtex}{}


%\end{IEEEbiography}


%\newpage



%\vfill


\end{document}